\documentclass[submission,copyright,creativecommons]{eptcs}
\providecommand{\event}{ICLP 2024} 

\usepackage{iftex}

\usepackage{tikz}
\usepgflibrary{shapes.geometric}
\usepackage{listings}
\lstdefinestyle{le}{
    commentstyle=\color{codegreen},
    keywordstyle=\color{magenta},
    basicstyle=\ttfamily\footnotesize,
    breakatwhitespace=false,         
    breaklines=true,                 
    captionpos=b,    
    keepspaces=true,                 
    numbers=left,                    
    numbersep=5pt,               
    showspaces=false,                
    showstringspaces=false,
    showtabs=false,                  
    tabsize=2,
    xleftmargin=10pt,
    escapechar=@,
    moredelim=[is][\color{blue}\bfseries]{~}{~},
    keywords={if,and,or,that}
}

\title{Mind the Gaps: Logical English, Prolog, and Multi-agent Systems for Autonomous Vehicles}
\author{Galileo Sartor
\institute{Department of Computer Science, Swansea University, United Kingdom}
\email{galileo.sartor@swansea.ac.uk}
\and
Adam Wyner
\institute{Department of Computer Science, Swansea University, United Kingdom}
\email{a.z.wyner@swansea.ac.uk}
\and
Giuseppe Contissa
\institute{Department of Legal Studies, University of Bologna, Italy}
\email{giuseppe.contissa@unibo.it}
}
\def\titlerunning{Mind the Gaps}
\def\authorrunning{Sartor, Wyner, and Contissa}
\begin{document}
\maketitle

\begin{abstract}
In this paper, we present a modular system for representing and reasoning with legal aspects of traffic rules for autonomous vehicles.
We focus on a subset of the United Kingdom's Highway Code (HC) related to junctions.
As human drivers and automated vehicles (AVs) will interact on the roads, especially in urban environments, we claim that an accessible, unitary, high-level computational model should exist and be applicable to both users.
Autonomous vehicles introduce a shift in liability that should not bring disadvantages or increased burden on human drivers.
We develop a system ``in silico'' of the model. 
The proposed system is built of three main components: a natural language interface, using Logical English, which encodes the rules; an internal representation of the rules in Prolog; and an multi-agent-based simulation environment, built in NetLogo. The three components interact: Logical English is translated into and out of Prolog (along with some support code); Prolog and NetLogo interface via predicates.
Such a modular approach enables the different components to carry different ``burdens'' in the overall system; it also allows swapping of modules.
Given NetLogo, we can visualize the effect of the modeled rules as well as validate the system with a simple dynamic running scenario.
Designated agents monitor the behaviour of the vehicles for compliance and record potential violations where they occur. The information on potential violations is then utilized by Validators, to determine whether the violation is punishable, differentiating between exceptions and cases.
\end{abstract}


\section{Introduction}

Research in autonomous vehicles is improving at a continuous pace. In a possible future, AVs and human agents will share a common space, such as city streets, where the AVs will have to interact with other road users in a predictible and understandable way.

Among the problems that need to be addressed specifically for this shared scenario is that of making the behaviour of AVs conform to the traffic laws \cite{DBLP:journals/ail/Prakken17}, in such a way as to not increase the burden on the other users of the roads.

For this reason we propose and present the development of rules from the UK Highway Code (traffic rules) that are modelled for both human and autonomous drivers.

The hypothesis we put forward is that with a combination of logic modelling and natural language we can obtain a representation of norms that can be directly used by both humans and autonomous agents, thus simplifying the inclusion of AVs on mixed roads.

The system we present does not make use of machine learning (e.g., image recognition) as input to the rule base. Rather we assume that the readings of the vehicle sensors are correct and fed into the rule base, so that the vehicle can reason on the input and determine which action to take accordingly. In future work, we might consider an integration between a rule-base and machine learning.

The system is designed to simulate multiple agents on the road, both human and autonomous and to detect violations in their behaviour. The violations are recorded by specific agents (monitors). The logs, with the information coming from the offending vehicle, can be used in a second moment to  perform legal reasoning on the violation and assign penalties when necessary.

The paper presents the background, outlining the main goals and issues in \ref{sec:background}, with particular attention on the issue of liability in \ref{sec:liability} and the existing projects/papers that explore the same or a similar space outlined in \ref{sec:SOA}.
The modular structure of the proposed system is then presented in \ref{sec:structure}, followed by the methodology for the design of the different component in \ref{sec:methodology}, with examples of the code.
The legal issues analyzed in the paper are further explored in the context of the proposed system in \ref{sec:lawsviolations}, specifically looking at how such a system could reason on the occurring violations, keeping in mind what the monitors logged, and also the reasoning the vehicles made at the time of the violation. This section will also look at what it means to violate a rule in the context of driving, what types of violations could occur, and their differing legal outcomes. We summarise and discuss future work in Section \ref{sec:summaryfuture}.

\section{Background}\label{sec:background}

Autonomous vehicles are going to share the road with human drivers, and other non autonomous agents. It is therefore crucial to ensure the behaviour of AVs is consistent with that of a \emph{good driver}, i.e. a human driver who follows the rules in a predictable manner.

This is even more important if we consider that, if well developed, these vehicles could be aware of their decision making, and could provide understandable explanations of the reasoning behind a certain action. As we will cover in the following sections, we consider this to be a crucial point, to reduce the burden on human agents who interact with the vehicle, and to reason with violations and reparations.

In developing the system, we should also consider the fact that as humans we make decisions also based on the possibility of incurring in violations and fines. There may be multiple reasons for this, and some may be legally valid, such as in the case of rules with exceptions, as will be described later. 
In the case of AVs however the general behavioural rules have to be determined at build time, given that there is no human agent involved in taking decision while driving. This question is made more complicated by the issue of liability.

\section{Liability}\label{sec:liability}

While autonomous vehicles are expected to drive more consistently with respect to the law and reduce accidents, violations of the law and accidents may still happen in certain situations. The question that arises is who should be held responsible for such violations and accidents - the driver, manufacturer, or the algorithm developer? Such issues are relevant for the general context of development of autonomous vehicles and guide how a system might be modeled. We develop issues below and specifically tie the concept of ``lawful reasonable agent'' to our implementation.

At the highest levels of automation, that is levels 4 and 5, according to the SAE definition \cite{on2021taxonomy}, the autonomous vehicle takes full control of all dynamic driving tasks. In these two levels of automation, the user is not expected to intervene when the automated driving system is on. Therefore, user’s liability is to be considered only when the ODD limits are exceeded (namely, the limits within which the driving automation system is designed to operate), or when users request the system to disengage the automated driving system. \footnote{This is consistent with the 2022 update to the Highway code, stating that \textit{While a self-driving vehicle is driving itself in a valid situation, you are not responsible for how it drives. You may turn your attention away from the road and you may also view content through the vehicle’s built-in infotainment apparatus, if available.}}

It should be noted that vehicles under level 4 or 5 may even be designed to be exclusively operated by technology, that is without user interfaces such as braking, accelerating, steering, etc. Therefore, such categories of vehicles do not contemplate at all the role of the human driver. Whenever user interfaces and controls are missing, the user will not be able to intervene in any of the dynamic driving tasks, and therefore cannot be considered in any way responsible for driving and subject to the related liabilities \cite{contissa2018liability}. 

Therefore, under levels 4 and 5, the burden of liability is mostly on manufacturers and/or programmers. They would be liable (1) when providing a defective or non-standard compliant tool that had a role in the causation of the accident, and (2) whenever the system fails to carry out the assigned task with a level of performance that is (at least) comparable to that reached by a human adopting due care under the same conditions.

This is linked to the idea that there would be a reasonable expectation that the autonomous vehicle performs an assigned task in a way that ensures the same level of safety that would be expected from a human performing the same task. Thus, it would seem appropriate to compare the autonomous vehicle's behaviour in carrying out an assigned task, with the behaviour that would be otherwise expected by the human driver \cite{patti2019autonomous}.

In this perspective, the concept of negligence would be central in evaluating the behaviour of the autonomous vehicle and assessing liabilities for manufacturers and programmers.

Negligence is `the omission to do something which a reasonable man, guided upon those considerations which ordinarily regulate human affairs, would do, or doing something which a prudent and reasonable man would not do' (Blythe v Birmingham Waterworks (1856) 11 Exch 781, at p 784). 

The tort of negligence usually requires the following elements: the existence of an injury, harm or damage; that the injurer owes \textit{a duty of care} to the victim; that the injurer has broken this duty of care (fault); that the damage (or injury) is a reasonably relevant consequence of the injurer’s carelessness (causal connection)\cite{bruggemeier2006common}. In the legal discourse, ``negligence'' denotes carelessness, neglect, or inattention, which are mental stances that can be ascribed only to human minds.

A driver of the vehicle has an asymmetrical duty of care toward pedestrians or other individuals in the vicinity.
The rules of 170-183 in part characterise how to meet this duty of care; broadly speaking, the driver should proceed defensively and cautiously, anticipating behaviours of others which might create circumstances in which the likelihood of an accident increases.

Clearly, the concept of negligence is linked to the idea of a human fault.
In contrast, liability for technological failures is usually evaluated and allocated on the grounds of product liability, which requires evidence of the following: an injury, harm or damage; a defective technology; and a causal connection between the damage (or injury) and the defect, namely that the former must be a reasonably relevant consequence of the latter. A technology may be considered defective if there is evidence of a design defect, a manufacturing defect, or a warning defect. Design defect, where the design is unreasonably unsafe, is the most relevant, and is usually determined by courts taking into account one of the following tests: the state of the art, the evidence of alternative design, or the reasonable expectations of users/consumers with regard to the function of the technology\cite{schebesta2017risk}. Key for our purposes is that negligence and product liability contrast with respect to duty-of-care and defective technology.

Yet, human and AI liability converge around performance: when a technology is used in substitution of a human, there is a reasonable expectation that the AI will perform an assigned task in a way that ensures the same level of safety that would be expected from a human performing the same task.

We reasonably assume that technology is presumed to have at least the same level of performance and safety as the human user. Thus, an AV ought to bear the same liability and duty of care as a human user. We propose that one code of conduct should rule both human and autonomous driver as a matter of fairness and equality on the road. This implies that we ought to be able to interrogate the automated vehicle on the same grounds as the human driver. To realise this, our modeling language should provide a unitary model, which yet allows for alternative means of realisation and data input.

\section{State of the Art}\label{sec:SOA}

In AI and Law, one of the main goals is to represent legal provisions as code \cite{Governatori2022-GOVTYO,BenchCapon2012}.
There are issues to address in order to obtain a good representation faithful to the source, and the main one is the presence of vagueness and open texture in the law \cite{DBLP:conf/dexaw/Bench-CaponV97}.
\cite{Bix2012-BIXDAO} discusses the legal context related to open textured concepts and defeasibility.
The natural language version of the HC has similar issues, that have already been discussed in the context of AVs with reference to natural language \cite{Irvine2023} and commonsense reasoning \cite{Kothawade2021}.

There have been multiple proposed approaches to addressing the issue of autonomous vehicles and the rules of the road.

An possible issue with the purely data-driven approaches is that there is a lack of well formed, diverse datasets \cite{Hammoud2022}, that are often biased towards accidents. Furthermore, they rely heavily on the interaction between autonomous vehicles \cite{Wang2022}, that are able to communicate, and act together as a swarm.

In this research we are attempting to address the interaction with and expectations of human agents on shared roads, where vehicles cannot rely on inter-vehicle communication. Further research is ongoing on how to apply data-driven predictive systems to mixed traffic (human and AVs).

We will focus on the rule-based approaches, and in particular those that enable further legal reasoning on the occurring events and actions.

In the research considered, the rules of the road (or a subset) have been modelled in higher order logic or temporal logic, to reason about the desired actions and concepts about the environment, in order to determine whether a certain action is valid. The agent in question can then take the desired action, and proceed with the movement.

One example of a similar representation is that of the RoTRA (Rules of The Road Advisor), \cite{DBLP:journals/corr/abs-2209-14035}, in which the rules are encoded in Prolog, and queried with respect to the state of the world (the context and beliefs), and the desired goal (intention).

Other projects have a more narrow focus on specific issues, such as determining the safe overtaking distances, with a formal model developed in Linear Temporal Logic, implemented in Isabelle/HOL, \cite{Rizaldi2017}.

The issue of encoding and reasoning with commonsense knowledge is not specific to the domain of autonomous vehicles, and is in fact a broader issue of knowledge based systems. In the context of driving, the analogy with human reasoning, and how modelling commonsense reasoning can help to develop reliable autonomous vehicles, is the topic of the AUTO-DISCERN (AUTOnomous DrivIng uSing CommonsEnse ReasoniNg) project\cite{Kothawade2021}.

\cite{DBLP:journals/ail/BhuiyanGBR24} presents an automatic compliance checking framework to assess AVs behaviour with respect to the traffic rules of Queensland, Australia. It considers issues related to open texture, exceptions, and conflicts amongst rules. 

In our research, we assume the driving environment has not been sterilised of human drivers, but rather includes them. Human understanding of and behaviour in the driving environment must be taken into consideration, which may go beyond the specification of the traffic rules (i.e., the interpretations) and require a unitary, transparent representation for both sorts of drivers, ensuring consistency.
We are also interested in the legal reasoning that occurs on the detected potential violations, and how the information from the vehicles can be used to reason about the specific scenario.

\section{Structure}\label{sec:structure}

The system presented is split in different, mostly independent modules, that each deal with one of the requirements and interact through minimal translation layers.\footnote{The full code of the system is available at \url{https://github.com/LyzardKing/mind-the-gap/tree/ICLP2024}}

\paragraph{The controlled natural language (CNL) module} is written in Logical English \cite{kowalskilogical}, syntactic sugar on Prolog, that enables to write rules and interact with the system in natural language. Using Logical English we can represent logic rules in natural language, that can be directly queried with a Prolog interpreter, or translated to Prolog for use by the autonomous vehicle.
\paragraph{The Logic rules module} is written in Prolog, and is mostly derived from the Logical English representation. The autonomous vehicle can reason with a Prolog interpreter, and use the result in determining its driving behaviour.
The Prolog output can be logged or converted back in natural language, saved in a human readable format, and can be used to check instances after the fact (scenarios and queries). This could be used in case of accidents or violations to determine why the vehicle took certain actions.
\paragraph{The simulation module} uses NetLogo, a multi-agent programmable modeling environment, where vehicles with different properties are  spawned, ad can move around on a predefined road grid. In addition to the basic movement, the vehicles can query the LE/Prolog rulebase to determine whether they are allowed to perform a certain action, or conversely, if they are prohibited.

In the following section the division of labour between the different components that was chosen for the system will be made clear.

\section{Methodology}\label{sec:methodology}

The development of the system started with the representation of norms in Logical English [Cite Mind the Gap]. Given the need for a simulation system, and the availability of different potential candidates, the idea was to keep the system modular, with the possibility to swap different components.
The current simulation uses NetLogo, but there is limited overlap in the components, mainly what is needed to convert data and I/O. The rules themselves can still be queried by LE/Prolog, and combined with other models.
The NetLogo simulation is derived from one of the examples made available in NetLogo\footnote{\url{https://ccl.northwestern.edu/netlogo/models/TrafficIntersection}}, and is then expanded through the use of a bridge to Prolog\footnote{\url{https://github.com/LyzardKing/NetProLogo}} that had previously been developed, and has been updated for the purpose of this project.
Vehicles in NetLogo are assigned different properties, and are spawned at one of the edges of the screen. They follow the road, and decide whether to turn at an intersection randomly. At the moment the system is not responsible for route planning, although in the future it might be added.
There are multiple types of vehicles, each with particular properties. Some of the properties are reflected in Prolog, while others are limited to the NetLogo environment.
Most properties and data come from sensors in the vehicle, that perceive the surrounding environment, as well as the properties inherent to the vehicle itself. The vehicles can see their surroundings, avoid other objects/agents, without accessing the legal norms. If we disable the Prolog section the vehicles can still drive, and act more like a swarm (Cite).
This behaviour may be very efficient on roads only used by autonomous vehicles, where the vehicles can communicate, and organize their actions accordingly. This is not the case on mixed roads, with both human and autonomous agents, pedestrians, bikes, emergency vehicles, and other potentially unpredictable agents.
Human drivers, while sharing the road with AVs, will have to be able to trust and understand the actions of the surrounding vehicles, while not necessarily knowing (or caring) if they are human or autonomous. In these circumstances, AVs will have to respect the same rules as their human counterparts, even if the actions are less optimized.

\subsection{Logic rules}

In the system the rules are represented in Logical English, in a way that is as isomorphic as possible to the original text. This makes the rules readable by humans, and could point to the possibility of having one simple corpus on which to write the rules, with them being automatically understandable and implemented by humans and autonomous agents.
The main goal here is to avoid repeating and maintaining multiple codebases, and to ensure the logic structure of the natural language version of the rules.
To assess the viability of such an approach the first rules modelled were those dealing with junctions (Rules 170-183 of the Using the Road section of the HC).
The first thing to note is that we are dealing with different types of morns, that may have different consideration when modelling: rules with a highlighted MUST, or MUST NOT, are those that are tied directly to laws (the Road Traffic Act 1988, The Traffic Signs Regulations and General Directions 2002), and deal with cases in which the driver is considered is guilty of an offence. We will visit these cases more in the next sections.
Most other rules deal not with explicit prohibitions/obligation, rather dictate what the behaviour of a good driver should be. In this case the terminology of the HC is very different, using words such as should, take extra care, look around, …
These terms are more nuanced, and while as humans we know how to deal with them, the same cannot be assumed of autonomous agents. For the representation to be adequate enough we may need in certain cases to add more information, and additional rules that form part of our commonsense reasoning.
Let us consider one of the modelled rules, rule 171, which states that:

\begin{quote}
    You MUST stop behind the line at a junction with a 'Stop' sign and a solid white line across the road. Wait for a safe gap in the traffic before you move off.
\end{quote}

This rule could be modelled by identifying the goal of the vehicle, entering the junction, and building the rule in Listing \ref{lst:junction}.
In the Logical English code, the word \textit{can} means \textit{has the permission to}, as used in the Highway Code.

\begin{figure}
    \centering
\begin{lstlisting}[style=le, caption=``Example rule in Logical English'', label={lst:junction}]
a vehicle can enter the junction if
the vehicle is of type ambulance
and it is not the case that
    the vehicle must give way to an other vehicle.

a vehicle can enter the junction if
the vehicle has green light
and it is not the case that
    the vehicle must give way to an other vehicle.
\end{lstlisting}
\end{figure}

In this case the rule expresses what should happen to the vehicle when approaching the junction. At first the vehicle should stop, since it is approaching a stop sign. Once it is next to the stop sign, the vehicle can query the system for its permission to enter the junction, and the second rule would be evaluated.
This is how the rules are currently modelled, and through further revisions they could be made more isomorphic depending on the specific needs.
The rules can be queried as is, by giving a scenario, a sample query that an AV could make, and could consequently show the solution in natural language, with all the steps that have been used to derive a certain conclusion (trace).
We are interested however in a more dynamic use of the rulebase, and for that we can rely on the previously mentioned agent simulation.

\subsection{Agent simulation}

The simulation is running currently in NetLogo, with the prolog extension to enable the agents to query the rulebase.
While currently there is only one Prolog process running, the single queries made by the vehicles are independent and isolated, to ensure that the queries are all atomic, and simulate a realistic scenario.

The rules in NetLogo are only those that pertain to the physical constraints, e.d. those actions the vehicles cannot physically make (e.g., occupying the same space of another vehicle). It is thus possible for the vehicles to drive without additional rules (in the same way as it is possible for human drivers to ignore the rules of the road). We then introduce the ``legal'' constraints, the Prolog rules. 

With the addition of the rules from the HC, the behaviour of the AVs becomes closer to what we would expect from human drivers.

\paragraph{Environment}

The simulated environment is very simple, consisting of three roads with two intersections. one of the intersections has a traffic light, while the other has stop signs. The intersections are spawned with their specific properties, and agents are generated independently starting from random road sections.

\begin{figure}
    \centering
    \includegraphics[width=0.8\textwidth,trim={7cm 7cm 7cm 7cm},clip]{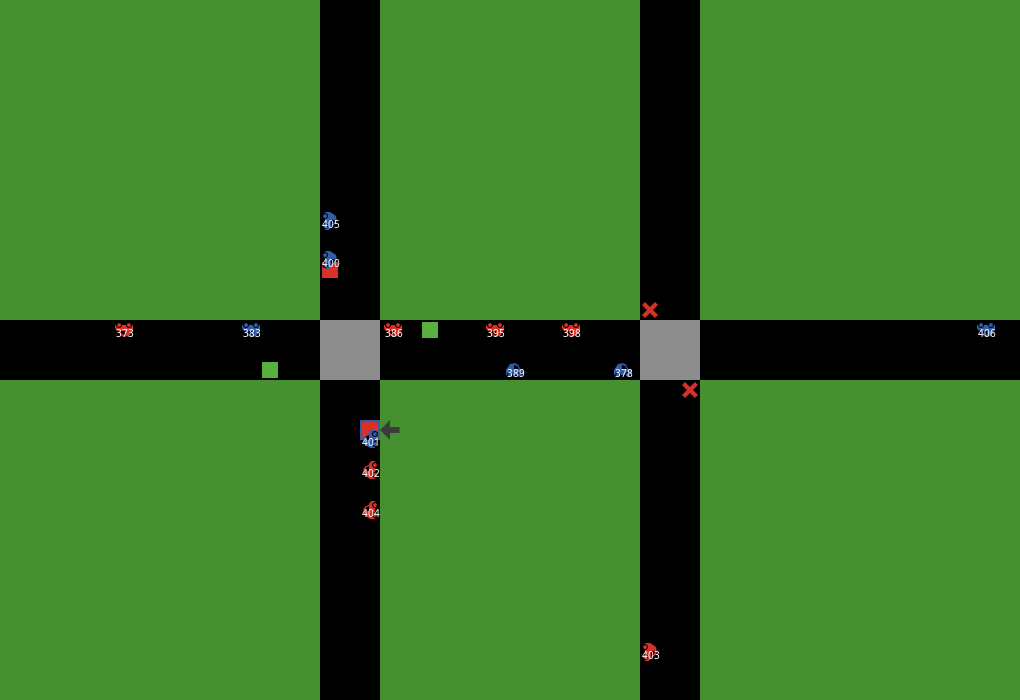}
    \caption{Simulated environment}
    \label{fig:simulated_env}
\end{figure}

\subsubsection{Agents}
In the simulation there are different agents, with different properties and goals:

\paragraph{Vehicles}
Vehicles are divided in two categories: cars, and emergency vehicles (ambulances). This is because rules may apply differently to different vehicle types.
At the moment in the simulation vehicles are divided in two categories: autonomous and human; as well as two types: cars and ambulances.

The different types of vehicles can be expanded, and share certain rules. In particular the main difference between the ambulances and cars is that the ambulances can violate certain rules of the HC, like crossing with a red light, so long as that doesn't directly cause an accident. To check this the vehicle uses the information about its surroundings to determine whether there is another vehicle close enough.

The cars are split in human and machine driven, with the main difference for now is the introduction of a number of delays and variations in the human behaviour. For example, a human driver may decide to go faster than the speed limit.

\paragraph{Pedestrians}
In the current simulation the pedestrian have a very basic behaviour, simply crossing when they get to a road. The only thing pedestrians will look at is if there is a car immediately approaching. This behaviour will be expanded upon in future development.

\paragraph{Monitors}

Monitors are the final agents that are active in the simulation. They focus on one section of road to see if they detect any vehicles which may have violated a traffic rule. The monitors have a narrow scope of vision and only access visible properties in the environment, e.g., cameras that recognize the license plate, speed and position of the vehicles; they  only react with respect to that information.
As such, monitors are purely reactive, rather than interactive. Moreover, they do not do any legal reasoning per se, which is why we only say they identify whether a vehicle may have violated a traffic rule a vehicle with its scope of coverage.

The rules that pertain to the monitors are modelled as in Listing \ref{lst:violations}. In the simulation, the monitor has vision of the traffic light, the vehicle position, and speed. When a vehicle enters the cone of vision of the monitor, the monitor gathers information about the traffic light and the vehicle speed; the monitor can detect whether the vehicle is moving or not, passing the predicate ``vehicle is stopped'' and the traffic light colour to Prolog rules. 
The Prolog rules used by the monitors then determine if a \textit{potential violation} occurred, i.e., if the vehicle is not stopped and the light is red.

\begin{figure}
\begin{lstlisting}[style=le, caption=``Example of the modeled violations", label={lst:violations}]
a vehicle potentially violates entering the junction if
the vehicle has red light
and it is not the case that
    the vehicle is stopped.
\end{lstlisting}
\end{figure}

As discussed later, information on potential violations is passed to a \textit{validator}, which may have additional information about the properties of a vehicle which can be taken (or not) to mitigate against issuance of a reparation. We say that if there is a potential violation and no mitigating circumstances, then there is a \textit{punishable violation}, which leads to a reparation.

\section{Violations and Penalties}\label{sec:lawsviolations}

While the HC is not in itself formally a legally binding document, it contains legal rules and references to the law, which indicate when violations arise and what is the correlated reparation (i.e., penalty to be paid; for our purposes, we use reparation and penalty interchangeably).

For instance: Failing to comply with traffic sign; Road Traffic Act 1988, s.36; £100; points 3. Also see Road Traffic Offences Act 1988 \url{https://www.legislation.gov.uk/ukpga/1988/53/part/II}. And the schedules with the penalties, for example, \url{https://www.legislation.gov.uk/ukpga/1988/53/schedule/2/part/I}. 
Our simulation must act and reason with respect to the violations and reparations.

In the simulation, vehicles can violate the HC rules in certain situations.
A human driver can independently decide if it is worth breaking a rule, depending on many factors, such as the probability of being caught, the probability of accidents, the change in time to reach the destination, and the amount of the potential fine.
Furthermore, whether or not a penalty is applied to an instance of a violation might depend on whether it is ``excusable' for one reason or another; that is, a violation is an exception to a norm, but some violations can themselves be exempted from penalty. For instance, a driver might be caught speeding, but not pay a penalty as they explain they were handling a medical emergency.
This general list of factors mostly applies to AVs as well with a caveat that there is no driver responsible (and liable) for deciding to break a rule, nor for the possible consequences.
As can be imagined, there are many factors with respect to which a norm is violated and conditions under which a penalty is or is not applied.

Given this, we work with a small domain to implement a vehicle's actions executed with respect to rules, whether the vehicle's action violates the rule, the detection of violations, consideration of mitigating circumstances, and the consequential penalties. As there are several rules, each related to actions; there can be correlated distinct violations, detections, mitigating circumstances, and penalties. In a sense, then, the \textit{actual} behaviour of a vehicle with respect to the rules of the road is compared to and evaluated against the \textit{ideal} behaviour as specified by the rules of the road. The \textit{ideal} behaviour is what the \textit{lawful reasonable agent} of Section \ref{sec:liability} would strive to achieve. Deviations of actual from ideal are noted and reasoned with further in terms of whether there were mitigating circumstances or not.

\subsection{Design}

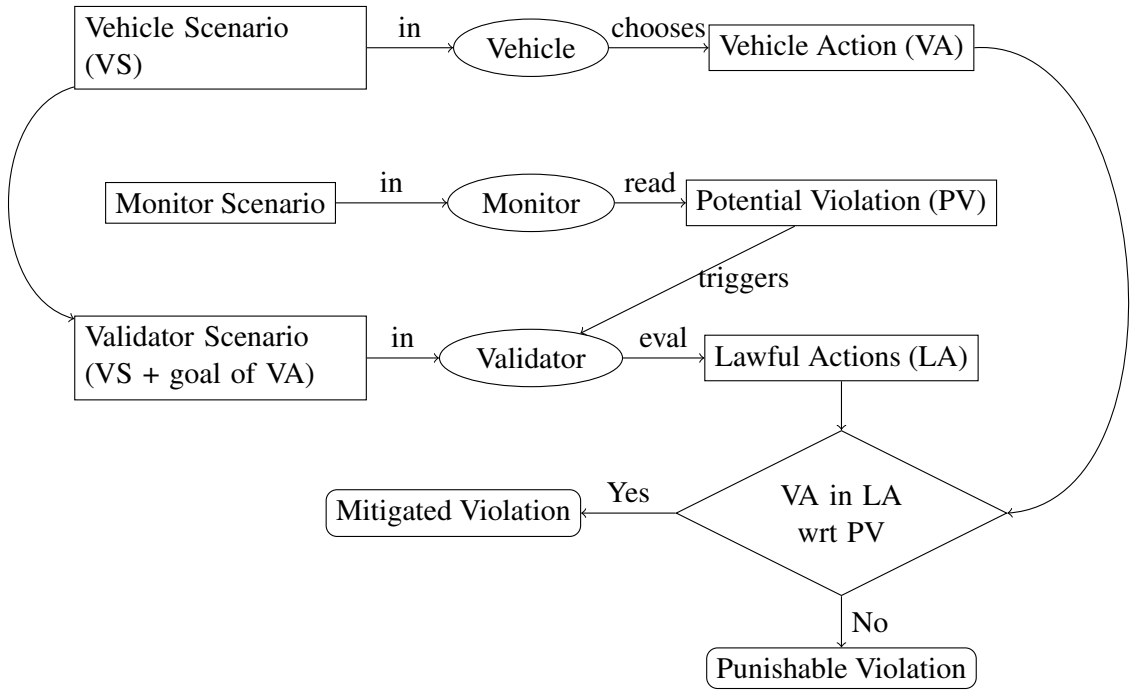
\begin{figure}
\resizebox{\columnwidth}{!}{%
\begin{tikzpicture}
\node[shape=rectangle, draw=black, text width=3.5cm] (vehicle-scenario) at (0, 4) {Vehicle Scenario (VS)};
\node[shape=rectangle, draw=black] (monitor-scenario) at (0, 2) {Monitor Scenario};
\node[shape=rectangle, draw=black, text width=3.5cm] (validator-scenario) at (0, 0) {Validator Scenario\\(VS + goal of VA)};
\node[shape=ellipse, draw=black] (vehicle) at (4, 4) {Vehicle};
\node[shape=ellipse, draw=black] (monitor) at (4, 2) {Monitor};
\node[shape=ellipse, draw=black] (validator) at (4, 0) {Validator};
\node[shape=rectangle, draw=black] (vehicle-action) at (8, 4) {Vehicle Action (VA)};
\node[shape=rectangle, draw=black] (potential-violation) at (8, 2) {Potential Violation (PV)};
\node[shape=rectangle, draw=black] (lawful-action) at (8, 0) {Lawful Actions (LA)};
\node[shape=diamond, draw=black, aspect=2, text width=2cm, align=center] (compare-violation) at (8, -2) {VA in LA\\wrt PV};
\node[shape=rectangle, draw=black, rounded corners] (punishable-violation) at (8, -4) {Punishable Violation};
\node[shape=rectangle, draw=black, rounded corners] (mitigated-violation) at (3, -2) {Mitigated Violation};
\path [->] (vehicle-scenario) edge node[above] {in} (vehicle);
\path [->] (vehicle-action) edge[bend left=90] node[above] {} (compare-violation);
\path [->] (monitor-scenario) edge node[above] {in} (monitor);
\path [->] (vehicle) edge node[above] {chooses} (vehicle-action);
\path [->] (monitor) edge node[above] {read} (potential-violation);
\path [->] (validator) edge node[above] {eval} (lawful-action);
\path [->] (validator-scenario) edge node[above] {in} (validator);
\path [->] (vehicle-scenario) edge[bend right=75] node {} (validator-scenario);
\path [->] (potential-violation) edge[] node[right] {triggers} (validator);
\path [->] (lawful-action) edge node {} (compare-violation);
\path [->] (compare-violation) edge node[right] {No} (punishable-violation);
\path [->] (compare-violation) edge node[above] {Yes} (mitigated-violation);
\end{tikzpicture}}
\caption{Action Execution with respect to Legal Rules}
\label{tikzpix01}
\end{figure}

Figure \ref{tikzpix01} is a graphic outline of the flow of information and reasoning. We start with Vehicle Scenario which is the state of the world within the scope of vision of the vehicle; it is the context in which the vehicle would execute an action (Vehicle Action). The Monitor is a reactive agent which is in charge of detecting a violation within its scope of vision which is the Monitor Scenario; they stand-in for cameras or the police. As a reactive agent, they record a Potential Violation, which remains to be validated with respect to the laws as indicated below. The Validator Scenario is  a hypothetical state of the world, one in which the Vehicle Scenario has been modified were the goal of the Vehicle Action to be attained. The Validator Scenario is used by the Validator to scope consideration of the Lawful Actions, which are those actions which are compliant with the laws in that Validator Scenario; in effect, we are given all those actions which, were they executed in the given Validator Scenario would be lawful. The Validator is triggered by an instance of a Potential Violation; it is used to evaluate whether the Potential Violation is indeed illegal or whether there might be mitigating circumstances. To move to this next step (VA in LA wrt PV), we consider whether the action that the vehicle executes (Vehicle Action) is amongst the Lawful Actions relative to the relevant Potential Violation, that is, whether the action has been caught by a monitor for a possible legal violation. If it is, then the violation is a Mitigated Violation; if not, then it is a Punishable Violation, which could require a penalty payment.

\begin{figure}
    \centering
    \begin{minipage}{.4\textwidth}
    \centering
\includegraphics[width=0.6\linewidth, trim={7cm 7cm 19cm 7cm},clip]{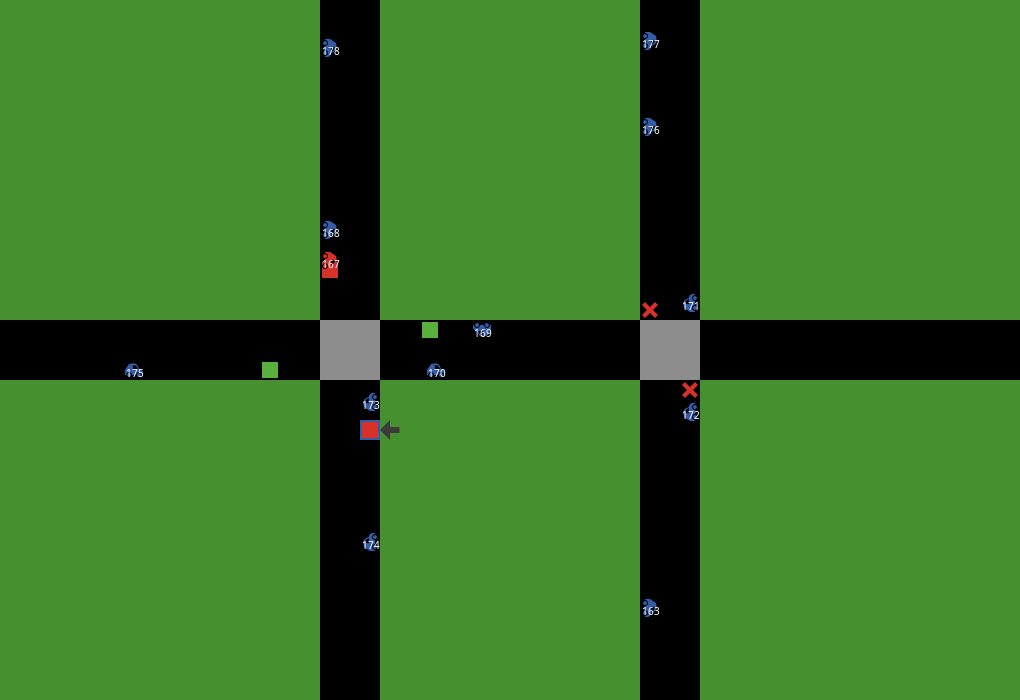}
    \caption{Example of violation detected by the monitor (the arrow)}
    \label{fig:violation}
    \end{minipage}
    \begin{minipage}{.4\textwidth}
    \centering
\includegraphics[width=0.6\linewidth, trim={7cm 7cm 19cm 7cm},clip]{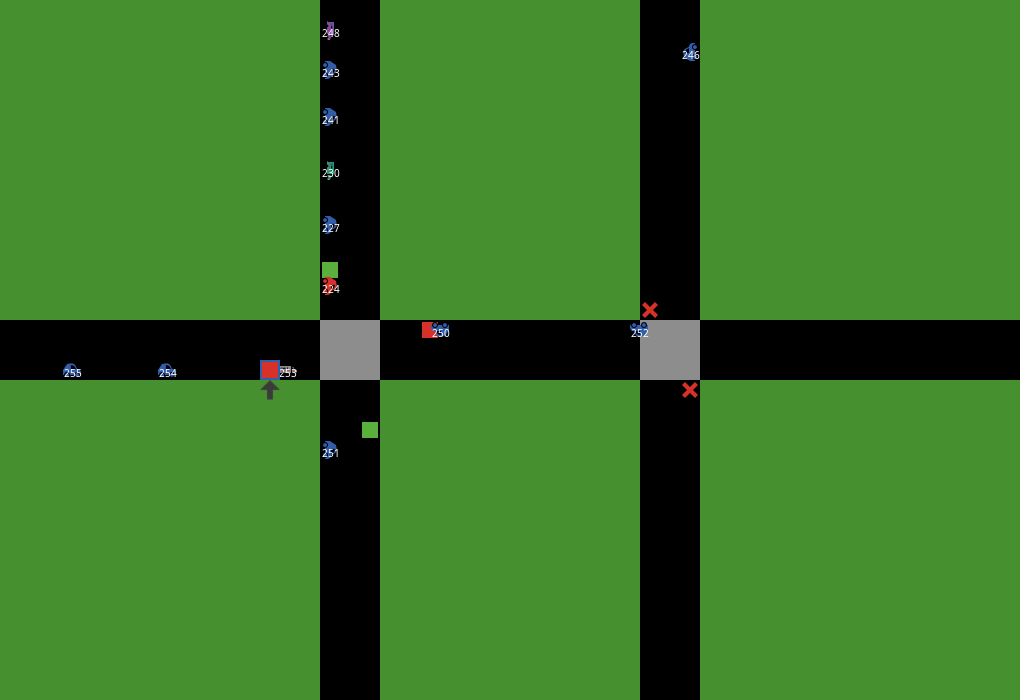}
    \caption{Example of an ``allowed'' violation, i.e., a permission}
    \label{fig:ambulance}
    \end{minipage}
\end{figure}

For example, in Figures \ref{fig:violation} and  \ref{fig:ambulance}, we have vehicles which enter the intersection against the red light. This introduces a Potential Violation in that whether the vehicle is penalised depends on whether or not it has mitigating circumstances relative to the law. An ordinary vehicle would raise a Punishable Violation in Figure \ref{fig:violation}, from which we would infer a correlated reparation (not shown). However, an ambulance would raise a Mitigated Violation, as in Figure \ref{fig:ambulance}, since as an ambulance it has a legitimate reason not to abide by the rule; consequently, no reparation can be inferred.

\subsection{Implementation}

Here we outline the implementation for each component of the design.

\paragraph{Scenario and Vehicle Action} Two possible Scenarios are in Listing \ref{lst:vehicle-scenario} and \ref{lst:ambulance-scenario}, which contain the goal of entering the junction.
The vehicle would execute an action (Vehicle Action), applying the rules in Listing \ref{lst:junction} to the Scenario, which provides rules for each of a car and an ambulance. Note that an ambulance does not need to abide by red lights, while a normal vehicle does.

\begin{figure}[ht]
\centering
\begin{minipage}{.4\textwidth}
\centering
\begin{lstlisting}[style=le, caption=``Normal vehicle Scenario'', label={lst:vehicle-scenario}]
scenario:
    173 is of type car.
    173 has red light.
goal:
    173 can enter the junction.
\end{lstlisting}
\end{minipage}
\begin{minipage}{.4\textwidth}
\centering
\begin{lstlisting}[style=le, caption=``Ambulance Scenario'', label={lst:ambulance-scenario}]
scenario:
    253 is of type ambulance.
    253 has red light.
goal:
    253 can enter the junction.
\end{lstlisting}
\end{minipage}
\end{figure}
\vspace{-1.0cm}
\paragraph{Monitor} The monitor can detect a violation, as discussed in relation to Listing \ref{lst:violations}, and records it.

\paragraph{Validator Scenario} is a hypothetical state of the world, one in which the Vehicle Scenario has possibly been modified where the goal would be realised by the Vehicle Action. In this instance, since the Potential Violation is related to the goal that the vehicles had in the Scenario and Vehicle Action above, the Validator Scenario is equivalent to the Vehicle Scenario. These need not be equivalent; for example, if the vehicle were caught speeding, though its goal were entering the junction.

\paragraph{Validator} The Validator uses the rules in Listing \ref{lst:add-validator-rules} with the Rules of the Road of Listing \ref{lst:junction} to determine the possible lawful actions, and compare them to the action which gives rise to the Potential Violation.

\paragraph{Comparing the Vehicle Actions, Legal Actions, and Potential Violations} The Listing \ref{lst:add-validator-rules} uses information about the recorded Potential Violation and whether the vehicle can execute the action (Listing \ref{lst:junction}). Where the vehicle can cross the red light and it is an ambulance, there are mitigating circumstances, so a violation is mitigated; where the vehicle cannot cross the red light, the violation is punishable.

\begin{figure}[ht]
\begin{lstlisting}[style=le,caption=``Comparing Ideal (Legal) Actions to Real (Violable) Actions'',label={lst:add-validator-rules}]
a vehicle punishably violates an action if
the vehicle potentially violates the action
and it is not the case that
    the vehicle can the action.

a vehicle mitigately violates an action if
the vehicle potentially violates the action
and the vehicle can the action.
\end{lstlisting}
\end{figure}

\section{Summary and Future Work}\label{sec:summaryfuture}

We have presented a modular framework for modeling autonomous vehicles that respect or violate the rules of the road and interact with other road users.
The modeled vehicles are designed in a way that makes their behaviour compatible with the behaviour of human agents, particularly with respect to violability.
The model includes violation detection and evaluation in a way that can take into account some different cases and exceptions.
As a modular system, components can be replaced with others, making it more complex, while maintaining the basic legal considerations and provisions. The overall function as outlined in Figure \ref{tikzpix01}, while the specific examples present a simplified instance.

In future work, we intend to report other aspects of the implementation and continue this research, expanding the rule-base and the simulation to closer map real world scenarios.
The legal reasoning component will be expanded, analyzing different natural language representations of rule priorities, exceptions, and the respective logic formulations.

The goal is to keep the rules modeled in a CNL as close as possible to the original source text. This may require tweaking parts of the existing code to ensure it is compliant with this requirement. In particular the definition of exceptions and rule hierarchy should be easily understandable and intuitive to human readers.

A possible line of research deals with the (partially) automated extraction of rules from the original source, so that it can be modeled as logic by an automated system and subsequently verified by human experts. This process would involve validating existing tools to automate the parsing of the text, and in particular their ability to keep the necessary level of consistency between the different rules.

Integration of machine learning approaches with legal reasoning would be an important avenue to explore, though how and where it integrates is an open question. While we would want to ``hard code'' from the HC, the overall system should have some flexibility to account for a variety of circumstances, e.g., open texture and commonsense reasoning.

The analysis of the violations, and the legal reasoning that occurs after the violations have been detected, will be expanded, to better define the rules that apply to AVs, and how the AV could behave in situations where a human driver would perhaps decide to violate a rule. As part of this, some integration with planning would be essential.

\section*{Acknowledgements}
The authors wish to thank Prof. Bob Kowalski, for his active leadership of the Logical English project, and for his encouragement to explore its applications to many different domains. We also thank Prof. Giovanni Sartor for supporting this work in the context of the H2020 ERC Project ``CompuLaw'' (G.A. 833647).


\bibliographystyle{eptcs}
\bibliography{bibliography}
\end{document}